\documentclass[1p,times]{elsarticle}
\usepackage[fontsize=11.3pt]{fontsize}
\PassOptionsToPackage{hyphens}{url}\usepackage{hyperref}
\usepackage{geometry}

\usepackage{amsmath,amssymb,amsfonts}
\usepackage{algorithmic}
\usepackage{graphicx}
\usepackage{textcomp}
\usepackage{xcolor}
\usepackage{caption}
\usepackage{subcaption}
\usepackage{float}
\usepackage{amsmath}
\usepackage{amssymb}
\usepackage{booktabs}
\usepackage{times}
\usepackage{epstopdf}
\usepackage{tikz}
\usetikzlibrary{positioning, decorations.pathreplacing, calc,fit,arrows.meta}

\usepackage{multirow}
\usepackage{tikz}
\usepackage{filecontents}
\usepackage[labelfont=bf]{caption}
\usepackage{breakcites}
\usepackage{lineno}
\usepackage[linesnumbered,ruled,vlined,onelanguage]{algorithm2e}
\usepackage{hyphenat}
\usepackage{subcaption}
\usepackage{etoolbox}
\usepackage{verbatim}
\usepackage{pdfpages}
\usepackage{etoolbox}
\makeatletter

\begin{document}
\makeatletter
\def\ps@pprintTitle{%
  \let\@oddhead\@empty
  \let\@evenhead\@empty
  \def\@oddfoot{}%
  \def\@evenfoot{}%
}
\makeatother
\begin{sloppypar}

\begin{frontmatter}

\title {GeoScatt-GNN: A Geometric Scattering Transform-Based Graph Neural Network Model for Ames Mutagenicity Prediction}                               


\cortext[cor]{Corresponding author}

\author[2]{Abdeljalil ZOUBIR\corref{cor}}
\ead{abdeljalil.zoubir@um6p.ma}

\author[2]{Badr MISSAOUI}
\ead{badr.missaoui@um6p.ma}

\address[2]{Moroccan Center For Game Theory, Rabat, UM6P, Morocco
}

\begin{abstract}

This paper tackles the pressing challenge of mutagenicity prediction by introducing three groundbreaking approaches. First, it showcases the superior performance of 2D scattering coefficients extracted from molecular images, compared to traditional molecular descriptors. Second, it presents a hybrid approach that combines geometric graph scattering (GGS), Graph Isomorphism Networks (GIN), and machine learning models, achieving strong results in mutagenicity prediction. Third, it introduces a novel graph neural network architecture, MOLG\(^3\)-SAGE, which integrates GGS node features into a fully connected graph structure, delivering outstanding predictive accuracy. Experimental results on the ZINC dataset demonstrate significant improvements, emphasizing the effectiveness of blending 2D and geometric scattering techniques with graph neural networks. This study illustrates the potential of GNNs and GGS for mutagenicity prediction, with broad implications for drug discovery and chemical safety assessment.

\end{abstract}

\begin{keyword}
Mutagenicity; Graph Neural Networks; Geometric Scattering Transform; Toxicity.
\end{keyword}

\end{frontmatter}

\section{Introduction}
In toxicological and pharmaceutical research, determining a compound's mutagenesis potential is critical to protecting public health and safety. This assessment is crucial in a number of domains, including medication development, environmental preservation, and conformity to regulations. The Ames test, a biological assay created by Bruce Ames in the 1970s, has long been considered the gold standard for determining chemicals capable of causing genetic changes \cite{ames1973carcinogens}. The test includes exposing specially modified Salmonella bacteria strains to the substance of interest and detecting whether or not it causes DNA alterations. Despite its great popularity and acceptability, the Ames test is not without flaws. It can occasionally produce erroneous results. The repeatability of the test between laboratories is not absolute, with a reproducibility of less than 100\% \cite{honma2019improvement}. These limitations, combined with the resource-intensive nature of the test and the ethical concerns associated with animal testing, have led the scientific community to seek alternative methods. The exponential increase in the number of new chemical entities requiring toxicological evaluation, coupled with the critical need for rapid screening in modern drug discovery pipelines, has accelerated the development of computational methods for predicting mutagenicity. These in silico methods hold the promise of faster, more cost-effective and potentially more accurate mutagenicity assessments that overcome the limitations of classical experimental methods while meeting the needs of today's research and regulatory environment.

As toxicology has evolved, the need for accurate mutagenicity assessment has increased, leading to the emergence of advanced computational models. While conventional machine learning (ML) \cite{feeney2023multiple} and deep learning (DL) \cite{li2023deepames} methods have shown potential, recent studies highlight the distinct advantage of graph neural networks (GNNs) \cite{li2021mutagenpred} for molecular analysis, including mutagenicity prediction. This advantage arises from the inherent graph structure of molecules, where atoms serve as nodes and bonds as edges, making GNNs particularly well suited to capturing molecular relationships.
Graph Neural Networks (GNNs) outperform typical machine learning models that rely on manually created features, as well as deep learning approaches that may overlook critical structural aspects. This ability to preserve molecular structure is crucial because it enables more detailed and nuanced examination of the molecular characteristics that determine mutagenicity.

Traditional machine learning models, while simpler, can struggle with the complexities of molecular data since they rely on predefined molecular descriptors and procedures. Although deep learning models, notably those based on convolutional neural networks (CNNs) \cite{van2024ampred} and recurrent neural networks (RNNs)\cite{winter2019learning}, have improved performance, they still struggle to capture the spatial and relational properties of molecules.
Accurate prediction of chemical toxicity is critical for ensuring the safety of new compounds. In this work, we present a novel approach that leverages the power of molecular graph representations based on scattering transform to predict zinc toxicity.

At the heart of our method is the Geometric Scattering Transform (GST), a powerful tool that can capture the intricate multi-scale structures within molecular graphs. We use GST to generate rich, informative embeddings that encode the essential features of each molecule.
These GST-derived embeddings serve two purposes. First, we use them as input to sophisticated machine learning models, which are then trained to directly predict the toxicity of molecules. Secondly, we use the embeddings to compute the similarity between molecules, forming a fully connected graph where the nodes features are the scattering coefficients and the edge weights reflect these similarities.

This graph-based representation is then processed using the GraphSAGE algorithm, which allows us to exploit the relational information between molecules to further improve the accuracy of our toxicity predictions. By combining the feature extraction capabilities of GST and the graph-based learning power of GraphSAGE, our approaches delivers state-of-the-art performance for zinc toxicity prediction.
The ability to accurately predict toxicity is a critical step in the development of new and safer chemicals. Our method, based on molecular graph representations and advanced machine learning techniques, represents a significant advance in this important area of research.

 \section{Related Work} \label{related}
 
Our work focuses on 2D and geometric scattering, graph neural networks (GNNs), and mutagenicity prediction. While earlier approaches have employed supervised machine learning models, CNN and GNNs to predict mutagenicity on the ZINC dataset, the application of scattering transform for feature extraction is relatively unexplored. In this paper, we close this gap by combining the geometric scattering transform with both supervised machine learning models like Lightgbm and GNNs like GraphSAGE. While the previous literature shows considerable gains in GNN-based mutagenicity prediction, our techniques improve prediction capabilities by including geometric scattering to gather multi-scale structure information. Here, we review key research, highlight the merits and limits of current approaches, and offer our work as a step forward in enhancing predictive accuracy for toxicology.

\subsection{Traditional Machine Learning Approaches in Mutagenicity Prediction}
Traditional machine learning (ML) models have been widely used in Ames mutagenicity prediction, generally using molecular fingerprints and specified descriptors to forecast toxicological effects. The Random Forest (RF) model is among the most widely utilized. Chu et al. \cite{chu2021machine} used RF with ECFP and FCFP features on a dataset of 5359 chemicals and obtained an AUC of 0.84 and an ACC of 0.79. Similarly, Venkatraman et al. \cite{venkatraman2021fp} used RF with PUBCHEM fingerprints (FP) on a larger dataset of 7950 chemicals, yielding an AUC of 0.87 and an ACC of 0.79. These researches established the usefulness of fingerprint-based descriptors for predicting mutagenicity by allowing models to extract key structural information from chemical substances.

Support Vector Machines (SVM) have also demonstrated excellent performance in this domain. Shinada et al. \cite{shinada2022optimizing} used SVM with ECFP4 fingerprints, chemical characteristics, and genotoxicity alerts to achieve an AUC of 0.93 across a dataset of 6512 compounds. The inclusion of diverse molecular descriptors enabled the SVM model to efficiently capture critical structural and chemical features related to mutagenicity, emphasizing the importance of feature selection in maximizing model performance.

Consensus models, which combine many methods, have also been investigated for improving mutagenicity prediction. Lou et al. \cite{lou2023chemical}  created a consensus model that included SVM, RF, XGBoost, Lightgbm, and GNN. Using characteristics such as MACCS keys, RDKit descriptors, and ECFP4 fingerprints, their technique achieved an AUC of 0.93 and an ACC of 0.87 on a dataset of 8576 chemicals. The success of consensus models stems from their ability to use the complementing qualities of many algorithms, resulting in increased robustness and predictive power across heterogeneous datasets.

In a 2024 study, Van Tran et al.\cite{van2024ampred} investigated the utilization of Lightgbm with RDKit 2D chemical descriptors with their AMPred-CNN model. The Lightgbm model has an AUC of 0.901 and an ACC of 0.827, indicating that classical machine learning methods can still yield high accuracy in mutagenicity prediction when paired with powerful molecular descriptors.

While these traditional ML models have been successful in many mutagenicity prediction tasks, they come with limitations. A key challenge is their reliance on predefined molecular descriptors such as ECFP, FCFP, and PUBCHEM FP, which may not fully capture the complex and non-linear relationships within molecular structures.

\subsection{Advances in Deep Learning for Predicting Mutagenicity
}
Deep learning algorithms have considerably improved mutagenicity prediction because of their capacity to learn complicated patterns from data. However, in some applications, deep neural networks (DNNs) are still used alongside molecular descriptors, which provide a structured representation of chemical substances. Kumar et al. \cite{kumar2021deep} used a DNN with alvaDesc molecular descriptors on a dataset of 4053 chemicals, attaining an AUC of 0.89 and an ACC of 0.84. In this scenario, while the DNN architecture allowed for the automatic learning of molecular feature connections, the alvaDesc descriptors still supplied a predefined structure to the data, making the learning work easier.

Similarly, Lui et al. \cite{lui2023mechanistic} used DNNs with Binary Morgan fingerprints to predict Ames mutagenicity. Their model, when applied to a dataset of 6380 chemicals, produced an AUC of 0.88 and an ACC of 0.78. The use of Morgan fingerprints as input features indicates that even deep learning models occasionally rely on feature engineering to capture crucial chemical attributes, rather than learning purely from raw data.

On the other hand, models such as the AMPred-CNN created by Van Tran et al. \cite{van2024ampred} used CNNs to learn directly from molecular images generated from SMILES strings. This method enabled the model to avoid typical feature engineering by utilizing CNNs' capacity to extract spatial and structural information from molecular representations autonomously. They also tested a Lightgbm model with RDKit 2D molecular descriptors and obtained an AUC of 0.901 and an ACC of 0.827, demonstrating the efficacy of combining classical machine learning and deep learning for mutagenicity prediction.

While deep learning models like as DNNs and CNNs have impressive potential for learning intricate chemical interactions, they are not without limits. There is a concern that models like the AMPred-CNN, which use CNNs to process molecular pictures, will overlook essential chemical structural information. Molecular pictures created from SMILES representations can occasionally oversimplify or obfuscate important chemical interactions like bond types, electron distributions, or three-dimensional spatial arrangements.

\subsection{Using Graph Neural Networks for Molecular Toxicity Prediction
}

Graph Neural Networks (GNNs) have emerged as an effective technique for predicting mutagenicity due to its ability to naturally model complicated chemical structures as graphs. Atoms are represented as nodes in these models, while bonds between them act as edges, allowing GNNs to successfully capture the intricate topological relationships seen within molecules. Xiong et al. \cite{xiong2021admetlab}  used GNNs for mutagenicity prediction with molecular graph data from ChEMBL, PubChem, OCHEM, and other sources. Their model, applied to 7575 molecules, obtained an AUC of 0.90 and an ACC of 0.81.

Further developments in Graph Convolutional Neural Networks (GCNNs) have expanded the range of mutagenicity prediction. Hung et al. \cite{hung2021qsar}  used a GCNN model on a dataset of 17,905 chemicals and obtained an AUC of 0.88 and an ACC of 0.85. Similarly, Li et al. \cite{li2021mutagenpred} created the MutagenPred-GCNN model, which combined molecular graphs with data-driven molecular fingerprints to achieve an AUC of 0.88 and an ACC of 0.81 on a smaller dataset of 6307 chemicals. These models demonstrate GCNNs' capacity to handle big datasets and learn complex chemical characteristics automatically from graph structure.

Guo et al. \cite{guo2022ligandformer} proposed a GNN model for predicting mutagenicity on a dataset of 7617 chemicals, with an AUC of 0.92. Their approach emphasized the model's strong interpretability, highlighting GNNs' ability to not only provide accurate predictions but also insights into molecular interactions important to mutagenicity. Wei et al. \cite{wei2022interpretable} also used GCNNs to analyze molecular graphs from ChEMBL, PubChem, DrugBank, and other publications. Their model achieved an AUC of 0.84 and an ACC of 0.84 on 7387 chemicals.

Although GNNs and GCNNs have distinct benefits in capturing topological links inside molecular structures, their usefulness is limited by the quality of the molecular graphs utilized as input. These graphs may not adequately capture crucial characteristics of molecular structure, such as 3D spatial configurations, electron distributions, or complicated bond dynamics. As a result, essential chemical features required for accurate mutagenicity predictions may be ignored, reducing the model's overall performance and generalizability.

To overcome these constraints, the Geometric Scattering Transform (GST) offers a more advanced method for obtaining multi-scale characteristics from molecule structures. Unlike standard graph representations, GST captures both local and global structural information, allowing for a more complete understanding of chemical interactions. This approach maintains the geometric structure of molecules, which is critical for accurately forecasting traits like mutagenicity.
By integrating GST with GNNs, we can improve the model's ability to detect intricate chemical properties that traditional graph-based approaches may overlook. GST introduces multi-resolution analysis, which allows GNNs to better understand molecular interactions at many scales, ranging from individual atom bonds to larger molecular substructures. This integration has the potential to outperform cutting-edge models by providing more precise and detailed representations of molecular characteristics, resulting in higher predicted accuracy in mutagenicity tasks.

\section{Preliminaries and Background}\label{Background}

\subsection{Wavelet Scattering Transform}\label{scatt}

The 2D Wavelet Scattering Transform (WST) is a powerful method for extracting robust, translation-invariant features from images while preserving high-frequency information often lost in traditional convolutional approaches. Structurally resembling a CNN, the WST provides a predefined, training-free alternative that excels on small datasets.

WST iteratively applies scaled and rotated wavelet convolutions, followed by modulus operations and smoothing, to generate hierarchical representations of images. At each layer, it performs wavelet transforms and nonlinearities; refer to \cite{lilly2008higher} for more background and references. Using Morlet wavelets, this method captures multi-scale and multi-orientation information, forming the foundation of the scattering transform.

\subsubsection{Morlet wavelets}

\begin{figure}[!th]
        \centering
        \includegraphics[width=0.40\columnwidth]{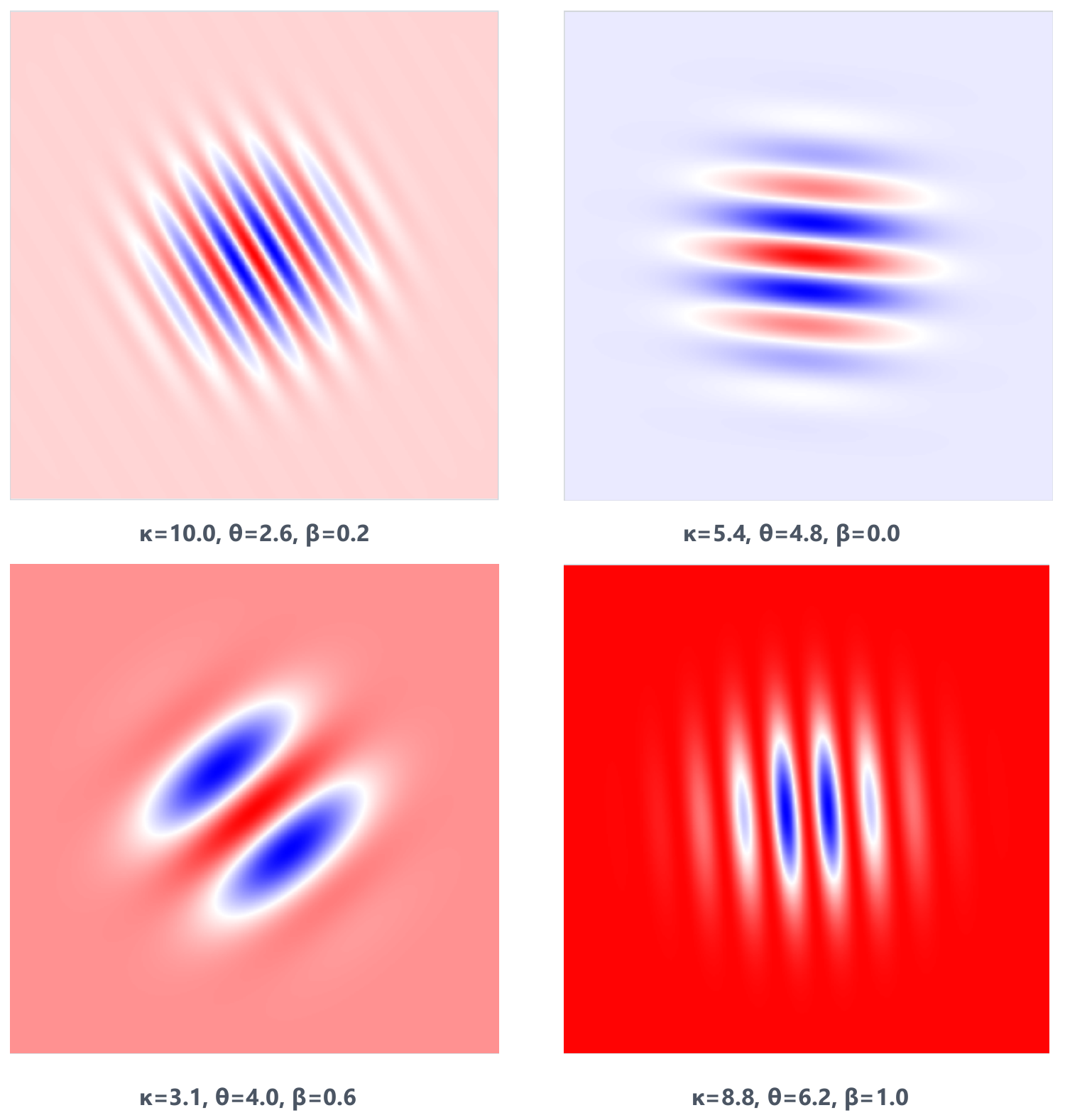}
    \caption{2D Morlet Wavelet visualization at various orientations and parameters}
    \label{morlet}
\end{figure}

The 2D Morlet wavelet \( \psi_{\theta}(x) \) for \( x \in \mathbb{R}^2 \) is defined as:  
\[
\psi_{\theta}(x) = e^{i k_{\theta} \cdot x} e^{-\frac{|x|^2}{2}} - \beta e^{-\frac{|x|^2}{2\sigma^2}},
\]  
where \( \theta \) is the orientation angle of the wavelet, \( \kappa \) is the wave vector magnitude, \( \beta \) is the DC (Direct Current) correction term, and \( \sigma \) is the width of the Gaussian envelope. We refer to Mallat \cite{mallat2008} for more background.\\
\noindent These parameters can be optimized for specific applications. Typically, \( \theta \) is sampled at multiple orientations for comprehensive directional analysis, \( \kappa \) is chosen based on the characteristic scale of the features of interest, and \( \beta \) is adjusted to control sensitivity to intensity variations in the analyzed signal. Figure \ref{morlet} plots the Morlet wavelet at various orientations and parameter settings.

\subsubsection{Scattering Transform}

Let $X$ be the signal to be analysed. A WST is implemented with a deep convolution network that iterates over traditional wavelet transform, nonlinear modulus, and averaging operators. \\
The zeroth-order scattering coefficient is obtained by averaging the input signal with a low-pass filter $\phi_J$:
\begin{equation}
S_0x = I \ast \phi_J,
\end{equation}
where $\phi_J(x_1,x_2) = 2^{-J}\phi(2^{-J}x_1,2^{-J}x_2)$ is a scaled version of the low-pass filter $\phi$.this later is the parent low-pass filter, typically chosen as a normalized 2D Gaussian function:
$$\phi(x_1,x_2) = \frac{1}{2\pi\sigma^2}\exp\left(-\frac{x_1^2 + x_2^2}{2\sigma^2}\right).$$

\noindent The first-order scattering coefficients are computed by convolving the input signal with oriented wavelets $\psi_{2^j,\theta}$, taking the complex modulus, and then averaging with $\phi_J$:
\begin{equation}
S_1(j,\theta)I = |I \ast \psi_{2^j,\theta}| \ast \phi_J,
\end{equation}
where $j$ represents the scale and $\theta$ the orientation. The wavelets at each scale are obtained by dilating the mother wavelet defined previously: $\psi_{2^j,\theta}(x_1,x_2) = 2^{-2j}\psi_\theta(2^{-j}x_1,2^{-j}x_2)$.

\noindent This process continues recursively to generate higher-order coefficients. The $m^{th}$ order scattering coefficients are computed by applying $m$ successive wavelet transforms and modulus operators, followed by a final averaging:
\begin{equation}\label{scatt_full_vector}
S_m(j_1,\theta_1,...,j_m,\theta_m)I = |||I \ast \psi_{2^{j_1},\theta_1}| \ast ... \ast \psi_{2^{j_m},\theta_m}| \ast \phi_J,
\end{equation}
where $j_1 < j_2 < ... < j_m < J$ ensures a coherent multi-scale analysis. The intermediate representations $U[j,\theta]I = |I \ast \psi_{2^j,\theta}|$ capture the modulus of wavelet coefficients before averaging, preserving high-frequency information that is essential for discrimination.

Several important mathematical aspects can be observed in the resulting scattering representation. The transform is locally translation invariant up to a scale of $2^J$, thanks to the final averaging with $\phi_J$. It is particularly resistant to tiny deformations, which is an important characteristic for pattern recognition applications. This stability results from a series of wavelet transforms and modulus operations that gradually increase invariance while conserving discriminative information.

In practical implementations in images, particularly molecular ones, the transform is often computed up to second order ($m \leq 2$), as higher-order coefficients have decreasing energy. The scattering coefficients at each order capture progressively complicated geometric patterns: first-order coefficients encode directional properties such as edges, whereas second-order coefficients record more intricate geometric connections and textural information. This hierarchical decomposition generates a rich, multi-scale representation of visual material that preserves fundamental properties while ensuring invariance to common changes.

The final representation is made up of concatenated scattering coefficients from all orders and scales as in (\ref{scatt_full_vector}), resulting in a vector that accurately describes the geometric and textural aspects of the original image. This representation has proven very useful for machine learning activities since it provides a consistent and meaningful set of features that may be utilized for classification, regression, and other predictions of chemical properties.

\subsection{Geometric Scattering on Graphs}\label{gst}
Geometric scattering on graphs extends the principles of the previous section on WST to graph-structured data \cite{Mallat2012}. This approach provides a rich, multi-scale representation of graphs that is invariant to vertex permutations, making it particularly useful for graph analysis, classification, and regression tasks \cite{gama2018diffusion, zou2020graph}.

In the field of graph-based machine learning, a fundamental challenge is developing algorithms that are invariant to graph isomorphism. This property ensures that the algorithm's output remains consistent regardless of how the vertices and edges of a graph are indexed, especially in molecular graph analysis. A common approach to achieve this invariance is through the use of summation operators, which act on signals defined on graphs. Let $G = (V, E)$ be a graph with vertex set $V$ and edge set $E$, and let $x = x_G$ be a signal defined on $G$.

While these simple summation-based features capture some graph properties, they do not fully represent the graph's structural information. To address this limitation, we turn to wavelet transforms on graphs. In this work,we associate each graph  an adjacency matrix \( W \), where \( W_{ij} = 1 \) if nodes \( i \) and \( j \) are connected, and 0 otherwise. The degree matrix \( D \) is a diagonal matrix where \( D_{ii} = \sum_{j} W_{ij} \), represents the degree of each node.

The graph Laplacian \( L = D - W \) serves as a fundamental operator, describing the difference between the degree and adjacency relationships. Alternatively, the normalized Laplacian:
\[
L_{\text{norm}} = I - D^{-1/2} W D^{-1/2},
\]
is often used to ensure stability and prevent biases due to nodes with high degrees.

The eigen decomposition of the normalized Laplacian gives us the graph Fourier basis:
\begin{equation}
L_{norm} = V \Lambda V^T,
\end{equation}
where $V$ is the matrix of eigenvectors and $\Lambda$ is a diagonal matrix of eigenvalues $\lambda$. These eigenvalues represent the frequencies in the graph spectral domain.

Graph wavelets are at the heart of geometric scattering, as they are designed to capture local and multi-scale aspects of graph-based signals. Several varieties of graph wavelets have been proposed. In our case studying graphs-based molecular analysis, we focus on two popular wavelets, diffusion wavelets and tight Hann wavelets.
\begin{itemize}
\item 
Diffusion wavelets \cite{coifman2006diffusion} are constructed from the matrix $T= \frac{1}{2}(I + D^{-1}W)$ raised to dyadic powers. Specifically, for $J \geq 1$, the diffusion wavelet filters at a scale $j$ are defined as
\begin{equation}
H_{0}= I+T,~~H_j = T^{2^{j-1}}(I - T^{2^{j-1}}),~~j\geq 1.
\end{equation}
Different powers of $P$ capture information between nodes at different scales, revealing electron exchange and chemical bonding patterns. 
In the ZINC dataset \cite{hansen2009benchmark}, diffusion wavelets effectively detect significant substructures, helping predictions of molecular attributes such as toxicity.

\item
In this work, we focus on tight Hann wavelets \cite{jiang2015tight} defined as:
\begin{equation}
\psi_j(\lambda_i) = h(\lambda_i - t_j), \quad i = 1, \ldots, N,
\end{equation}
where $\lambda_i$ represents the eigenvalue i in the graph Laplacian and $t_j = (j+1) e_{max} / (J + 1 - R)$ allowing for a flexible and energy-preserving representation of the graph structure, and
the Hann kernel function $h$ defined as:
 \begin{equation}
h(x) = 0.5 + 0.5 \cos\left(2\pi \frac{J + 1 - R}{R e_{\text{max}}} x + 0.5\right),
\end{equation}
where $J$ is the number of scales, $R$ is a scaling factor, and $e_{\text{max}}$ is the upper bound on the spectrum. The tight Hann wavelets are then constructed using the eigenvectors and eigenvalues of the normalized Laplacian.
It's important to note that $\psi_j$ is applied element-wise to each eigenvalue $\lambda_i$. This results in a vector of wavelet coefficients for each scale $j$:
\begin{equation}
\boldsymbol{\psi}_j = [\psi_j(\lambda_1), \psi_j(\lambda_2), \ldots, \psi_j(\lambda_N)].
\end{equation}

The wavelet transform on graphs is implemented through wavelet operators $H_j$. These operators are defined in terms of the graph Fourier basis and the wavelet functions:
\begin{equation}
H_j = V \text{diag}(\boldsymbol{\psi}_j) V^T,
\end{equation}
where $V$ is the matrix of eigenvectors of the graph Laplacian, and $\text{diag}(\boldsymbol{\psi}_j)$ creates a diagonal matrix with the elements of $\boldsymbol{\psi}_j$ on its diagonal.

\end{itemize}

Let \( X \in \mathbb{R}^{N \times F} \) be the graph signal where \( N \) is the number of nodes and \( F \) is the dimension of the feature vector at each node. Given these graph wavelets, the zeroth and $m$-order scattering coefficients are defined by:
\begin{eqnarray}
S_0X &=& \langle X, U \rangle = \frac{1}{N} \sum_{i=1}^{N} X_i, \xspace   X_i \in \mathbb{R}^F \\
S_m X(j_1, \dots, j_m) &=& \left\langle \left\vert H_{j_m}\left(\left\vert H_{j_{m-1}}\left(\cdots \left\vert H_{j_1}X \right\vert \right)\right\vert\right)\right\vert, U \right\rangle , ~~ j\geq 1.
\end{eqnarray}

\begin{figure*}[!th]
        
        \includegraphics[width=\textwidth]{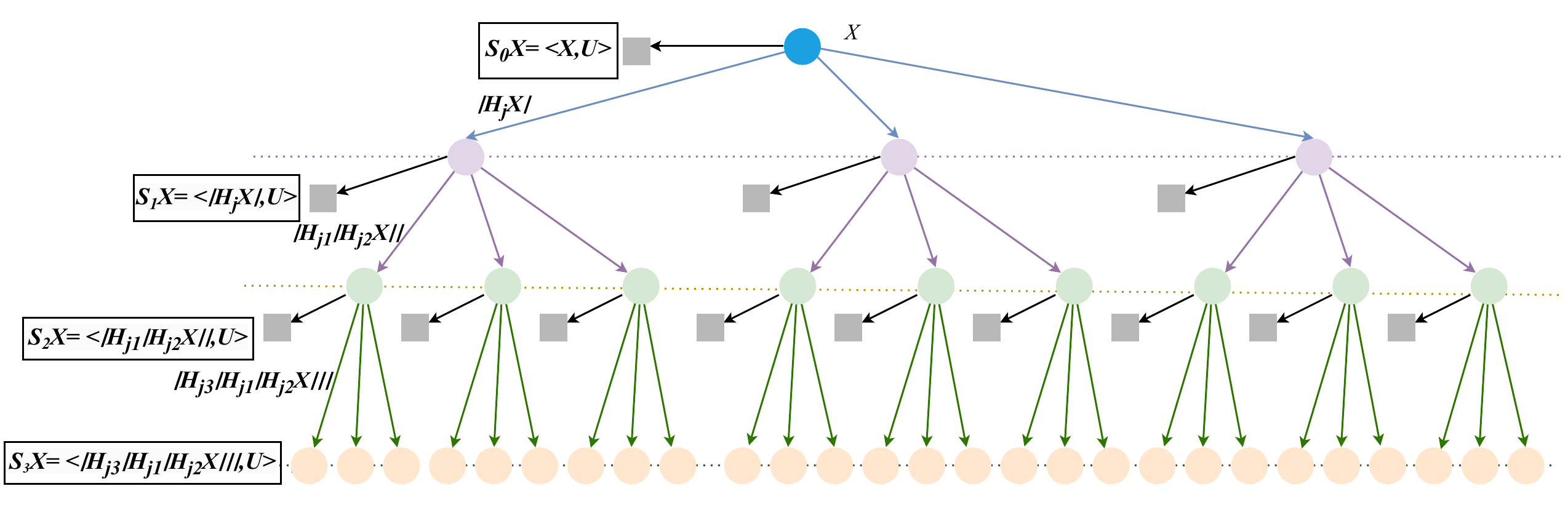}
    \caption{Graph Scattering Transform with \( J=3 \) and \( L=3 \) for multiscale signal decomposition.}
    \label{figSg}
\end{figure*}

The graph scattering transform is then defined by concatenating all of these coefficients: 
 \[
 S X = [S_0X, S_1X, S_2X, \dots, S_mX].
 \]
 Figure \ref{figSg} illustrates the hierarchical process of the graph scattering transform, where a graph signal 
\( X \) is filtered through \( J=3 \) scales using graph wavelets and processed over 
\( L=3 \) layers. At each layer, non-linearities and aggregation operators capture multi-scale hierarchical features of the signal.

\subsection{Graph Neural Network}\label{GNN}
Graph Neural Networks \cite{scarselli2008graph} are deep learning models that analyze graph-structured data. They work by iteratively updating node representations via message transmission among surrounding nodes. The basic premise is that each node's attributes are updated using both its own qualities and aggregated information from its neighbors. This method enables GNNs to capture both node-level features and structural information embedded in the graph topology.

In general, the message passing framework for GNNs can be expressed as:

\begin{equation}
h_v^{(k)} = \text{UPDATE}^{(k)} \left(h_v^{(k-1)}, \text{AGGREGATE}^{(k)}\left({h_u^{(k-1)} : u \in \mathcal{N}(v)}\right)\right),
\end{equation}
where $h_v^{(k)}$ represents the feature vector of node $v$ at layer $k$, $\mathcal{N}(v)$ denotes the neighbors of node $v$, and AGGREGATE and UPDATE are learnable functions.

\subsubsection{GraphSAGE}
GraphSAGE \cite{hamilton2017inductive} is an inductive learning framework for GNNs that enables generating embeddings for previously unseen nodes. The key innovation of GraphSAGE lies in its neighborhood sampling strategy and aggregation functions.
The message passing in GraphSAGE can be formulated as:
\begin{equation}
h_v^{(k)} = \sigma\left(W^k \cdot \text{CONCAT}\left(h_v^{(k-1)}, \text{AGG}_k({h_u^{(k-1)}, \forall u \in \mathcal{N}(v)})\right)\right),
\end{equation}
where $\text{AGG}_k$ can be any differentiable aggregator function (e.g., mean, max, or LSTM), $W^k$ is a learnable weight matrix, and $\sigma$ is a nonlinear activation function. The CONCAT operation ensures that the model preserves the target node's own features alongside the neighborhood information.
\subsubsection{Graph Isomorphism Network (GIN)}
GIN \cite{kim2020understanding} is designed to be as powerful as the Weisfeiler-Lehman graph isomorphism test in distinguishing graph structures. It achieves this by using a simple but powerful update function that maintains injective aggregation of multisets.
The GIN layer update rule is defined as:
\begin{equation}
h_v^{(k)} = \text{MLP}^{(k)}\left((1 + \epsilon^{(k)}) \cdot h_v^{(k-1)} + \sum_{u \in \mathcal{N}(v)} h_u^{(k-1)}\right),
\end{equation}
where $\epsilon^{(k)}$ is either a learnable parameter or fixed to zero, and MLP is a multi-layer perceptron. The $(1 + \epsilon^{(k)})$ term helps the model to distinguish central nodes from their neighbors, while the summation provides a simple yet powerful aggregation scheme that preserves multiset properties.
The key distinction of GIN is its proven theoretical capacity to capture structural information with maximal discriminative power among GNNs, making it particularly effective for graph-level tasks where isomorphism testing is important.
\section{Methods}\label{molecular}

\subsection{Dataset Preparation}

We utilized the well-known mutagenicity dataset \url{http://doc.ml.tu-berlin.de/toxbenchmark/} compiled by Hansen et al.\cite{hansen2009benchmark}, which originally contained 6,512 compounds, to assess and compare the effectiveness of our approach against previous methods. To preprocess and harmonize the molecular data, we stripped explicit hydrogens, detached and discarded any metal ions, and kept only the largest fragments of the molecules. To handle duplicate entries with identical canonical SMILES strings but differing mutagenicity outcomes, we applied the clear evidence rule. According to this rule, when two Ames test results conflict, preference is given to the positive outcome. After this filtering process, a refined set of 6,277 compounds was obtained, comprising 3,388 mutagenic and 2,889 non-mutagenic compounds. We then divided the dataset randomly, allocating 80\% for training and validation purposes and the remaining 20\% for testing.

\subsection{Scattering Transform for Hierarchical Molecular Representation Learning}

\subsubsection{Geometric Scattering-Based Molecular Featurization}

Molecular graphs are a strong tool for representing the structure of chemical compounds by storing atoms as nodes and bonds as edges. Each molecular structure is turned into a graph $G = (V, E)$, where $V$ represents the set of atoms and $E$ the chemical bonds between them. Individual atom parameters such as atomic number, formal charge, aromaticity, hybridization, and valence are incorporated as node features, whereas bond types contribute to the adjacency matrix, which defines the graph's connectedness. These graph-based models are critical for reflecting both local surroundings (such as aromatic rings or functional groups) and long-range interactions inside the molecule.

\begin{figure*}[!t]
        
        \includegraphics[width=\textwidth]{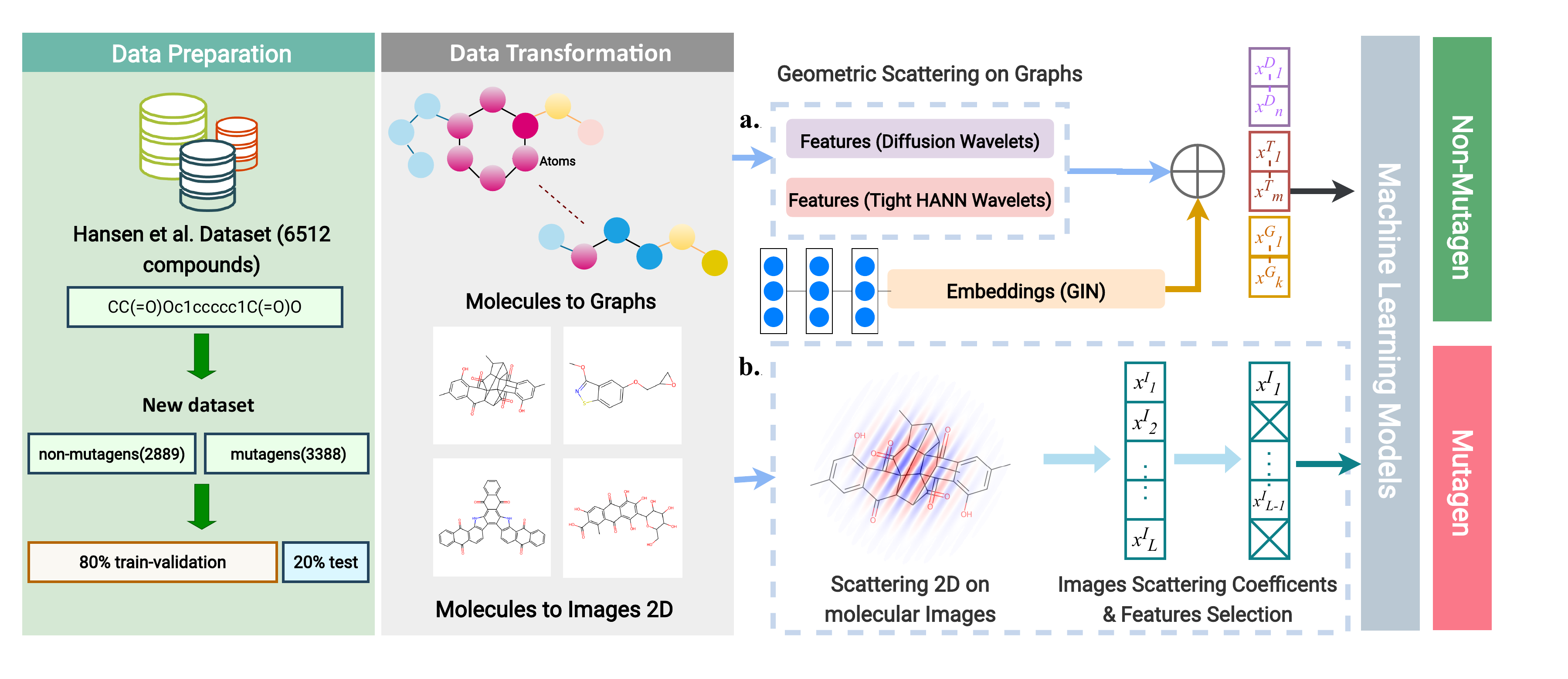}
    \caption{Multi-Modal Pipeline for Mutagenicity Prediction Using Molecule Representations as Graphs and Images. The pipeline utilizes the Hansen et al. dataset of 6,512 compounds divided into mutagens and non-mutagens. Molecules are transformed into graph representations for geometric scattering (a) using Diffusion and Tight HANN wavelets, along with embeddings from a Graph Isomorphism Network (GIN), and into 2D molecular images for scattering coefficient extraction (b). These features are fused and fed into machine learning models to classify compounds as mutagenic or non-mutagenic.}
    \label{fig1g}
\end{figure*}

To generate expressive embeddings, as shown in Figure\ref{fig1g}a,  geometric scattering transforms are applied to these molecular graphs.These transforms use wavelet-based filters to capture patterns at various scales and layers, increasing the feature space for subsequent tasks such as molecular property prediction. In our study, we use two complimentary wavelet-based transforms: Tight Hann Frame Wavelets and Diffusion Wavelets. Each technique provides a distinct view of the graph, guaranteeing that both local atomic surroundings and global molecule interactions are reflected in the embedding.

The Tight Hann wavelet transform \cite{jiang2015tight} is a spectral graph scattering technique that relies on localized filtering in both the node and frequency domains. The Hann window function is used to create wavelets that balance spectral localization and leakage. This approach detects small-scale patterns, such as aromatic rings and functional motifs, which are necessary for molecular property prediction.

In our approach, as illustrated in table \ref{gstv}, we use one layer  of the tight Hann wavelet transform, driven by the compact size of molecular graphs. For smaller graphs, a single layer strikes a balance between computational efficiency and the ability to extract meaningful patterns, such as functional groups and aromatic rings. To further enrich the structural representation, we apply the wavelet across multiple scales. After experimenting with various combinations of layers and scale values, we found that using three scales (\( j=3 \)) provided the best results. This multi-scale capability ensures that both local motifs and short-range dependencies are captured in the embeddings, making it particularly effective for datasets like ZINC in our case, where molecules exhibit diverse structural patterns.

The Diffusion wavelet's structure \cite{coifman2006diffusion} allows for the extraction of both local and global information, allowing the model to identify electron exchange patterns, chemical bonds, and functional connections at multiple scales. At this step, we use the diffusion wavelet transform with three layers. Each layer represents a bigger neighborhood interaction: the first encodes local atomic interactions, the second catches mid-range dependencies, and the third discloses global molecule features. This multi-scale propagation preserves and amplifies structurally significant substructures, such as rings, chains, and long-range relationships, in the final embeddings.

This dual-wavelet technique is especially useful for molecular property prediction tasks, such as toxicity evaluation and activity prediction, because it takes advantage of both localized motifs and global molecular interactions. By combining the strengths of spectral filtering (Hann wavelets) and random-walk-based propagation (diffusion wavelets), the geometric scattering framework produces durable and expressive embeddings suited for a wide range of downstream machine learning tasks.

\begin{table}[!ht]
\small
\centering

\begin{minipage}{0.8\linewidth} 
\centering
\renewcommand{\arraystretch}{1.5}
\caption{Hyperparameter values used in Geometric Scattering.}
\label{gstv}
\scalebox{0.85}{
\begin{tabular}{*2l}
    \toprule
    Hyperparameter & Values \\
    \toprule
    No. Tight HANN Layers & 1 \\
    No. Tight HANN Scale \( J \) & 4 \\
    No. Tight HANN coefficients & 7 \\
    No. Diffusion Wavelet Layers & 3 \\
    No. Diffusion Wavelet Scales & 4 \\
    No. Diffusion Wavelet Coefficients & 588\\
\end{tabular}
}
\end{minipage}

\end{table}

\subsubsection{Molecular featurization using scattering 2D}

The 2DWST uses multi-scale wavelet convolutions to successfully capture both local and global molecular characteristics. RDKit converts molecular graphs into 2D images while maintaining important substructures such as aromatic rings, functional groups, and bond configurations. 2DWST has both translation invariance and deformation robustness, making it ideal for studying the variety of molecular structures present in the dataset ZINC.

The 2DWST consists of Morlet wavelet convolutions followed by non-linear modulus operations. First-order coefficients capture basic chemical properties such as bonds and edges (Figure \ref{fig1g}b), but second-order coefficients encode more sophisticated interactions like ring connectedness. Higher orders provide less energy and are frequently shortened for efficiency. After extensive experimentation, The chosen combination of \( J=9 \) scales and \( L=8 \) orientations ensures that both local and global structures are fully represented while minimizing overhead. This is critical for datasets such as ZINC, where molecular variety necessitates a model that is sensitive to small chemical differences while remaining computationally tractable.

After applying the 2DWST to molecular pictures, each image produces an 11,681-dimensional feature vector. This high-dimensional representation, while dense in information, poses considerable processing hurdles. To address this, we used the Chi-squared ($\chi^2$) feature selection method to determine the most important factors determining mutagenicity. We tested different feature subset sizes, including 1000, 2000, 4000, 6000, and 8000. The results \ref{res} show that a subset of 4000 characteristics is ideal for balancing computational efficiency with predictive effectiveness.

\subsection{Molecular Graph-of-Graphs with Geometric Scattering and SAGE (\textbf{MOLG\(^3\)-SAGE})}\label{gsts}
In this study, we propose an alternative method MolG³-SAGE (Molecular Graph-of-Graphs with Scattering and GraphSAGE), a novel approach for molecular structure classification that models molecules as hierarchical, connected graphs. Each molecule is represented as a node in a fully connected meta-graph, with the nodes being individual molecular graphs made up of atomic and bonding information. This graph-of-graphs structure allows us to encode both intra- and intermolecular interactions, resulting in more sophisticated and hierarchical data representations.

\begin{figure*}[!ht]
        \centering
        \includegraphics[width=\textwidth]{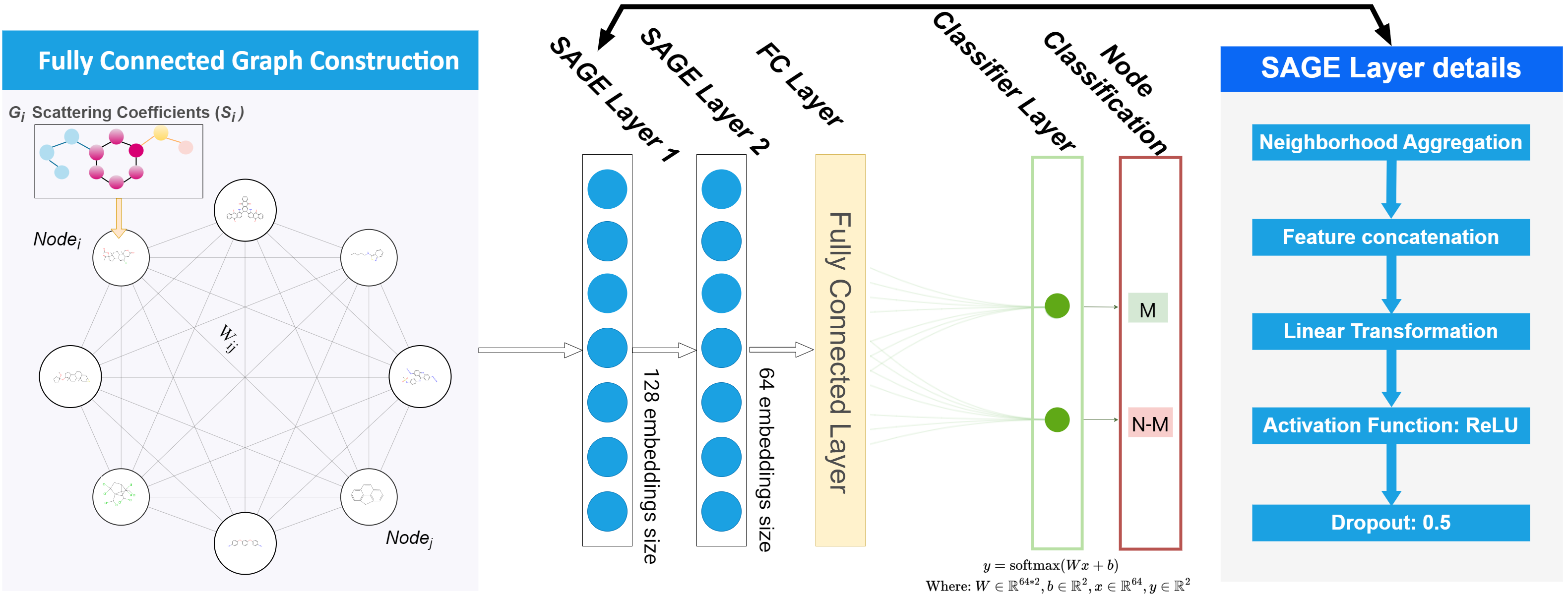}
    \caption{Fully Connected Layer Architecture for Binary Mutagenicity Classification in MolG³-SAGE Framework}
    \label{fig2g}
\end{figure*}

The construction of the meta-graph as shown in figure\ref{fig2g} involves creating weighted edges between molecular nodes based on their geometric scattering feature similarities. For each pair of molecules $i$ and $j$, we first compute their scattering embeddings, denoted as $s_i$ and $s_j$ respectively. The similarity between these embeddings is captured through cosine similarity, which is then transformed into a normalized weight $W_{ij}$. Let $S = \{s_1, ..., s_n\}$ be the set of scattering embeddings for $n$ molecules. The edge weight $W_{ij}$ between molecules $i$ and $j$ is defined as

\begin{equation}
\text{cos\_sim}(i,j) = \frac{\langle s_i, s_j \rangle}{\|s_i\| \|s_j\|},
\end{equation}

where $\langle\cdot,\cdot\rangle$ denotes the inner product and $\|\cdot\|$ represents the $L_2$ norm. To ensure non-negative weights bounded between $[0,1]$, we normalize the similarity:

\begin{equation}
\text{sim\_norm}(i,j) = \frac{\text{cos\_sim}(i,j) + 1}{2}.
\end{equation}

\noindent The similarity is then converted to a distance metric:

\begin{equation}
d(i,j) = 1 - \text{sim\_norm}(i,j).
\end{equation}

Finally, we apply a Gaussian diffusion kernel \cite{coifman2006diffusion} to obtain the final edge weights:

\begin{equation}
W_{ij} = \frac{\exp(-d(i,j)^2/(2\sigma^2))}{\max_{k,l} \exp(-d(k,l)^2/(2\sigma^2))},
\end{equation}

where $\sigma$ is chosen as the standard deviation of the distance matrix to ensure appropriate scaling of the kernel, and the denominator normalizes the weights to $[0,1]$. This formulation ensures that the edge weights decay exponentially with molecular dissimilarity while preserving local structure through the Gaussian kernel.

The resulting weighted graph $G = (V, E, W)$ consists of $|V| = n$ nodes representing individual molecules, with $E$ containing all possible edges between nodes, and $W$ containing the computed edge weights. Each node $v_i \in V$ maintains its original molecular graph structure $G_i = (V_i, E_i)$ representing atomic connections, thus creating a hierarchical graph representation. This dual-level structure allows MolG$^3$-SAGE to simultaneously capture both local atomic interactions through the molecular graphs and global molecular relationships through the weighted meta-graph structure.

This hierarchical structure enables MolG$^3$-SAGE to effectively capture both local atomic interactions and global molecular similarities, learn from the relationship between molecular structures, and leverage both geometric and topological information for enhanced message passing during graph neural network training.
 As illustrated in Table \ref{molg3} and Figure \ref{fig2g}, MolG³-SAGE's design includes two sequential GraphSAGE layers \cite{hamilton2017inductive} and a classification layer. Each GraphSAGE layer employs a neighborhood aggregation technique, which generates node representations by sampling and aggregating information from local graph neighborhoods. The first layer converts the initial molecular features into an intermediate representation, which is then refined further to produce a final embedding space. To avoid overfitting, a dropout mechanism is used between layers, and the ReLU activation function is used to add nonlinearity to the model.

Our method uses a message-passing framework in which each node accumulates information from its neighbors using a mean-pooling process. The aggregated neighborhood features are added to the node's own features before being converted using a learned linear projection. This approach enables the model to recognize both local structural patterns and global chemical characteristics. Edge weights in the meta-graph are calculated using chemical similarity metrics, resulting in a weighted connectivity structure that reflects the interactions between distinct molecules.

To minimize cross-entropy loss, the Adam optimizer is used during the training process, along with weight decay regularization. We use an early stopping mechanism based on validation loss to prevent overfitting and model checkpointing to keep the best-performing configuration. The model's performance is evaluated using traditional binary classification measures, with a focus on the area under the receiver operating characteristic curve (ROC-AUC), which is robust to class imbalance.
\begin{table}[!ht]
\small
\centering

\begin{minipage}{0.8\linewidth} 
\centering
\renewcommand{\arraystretch}{1.5}
\caption{Hyperparameter values used in the MolG³-SAGE model for mutagenicity classification.}

\label{molg3}
\scalebox{0.85}{
\begin{tabular}{*2l}
    \toprule
    Hyperparameter & Values \\
    \toprule
    Number of Layers & 2 \\
    Input Feature Dimension & 595 \\
    Hidden Features & 128 \\
    Embedding Size & 64 \\
    Dropout Rate & 0.5 \\
    Learning Rate & 0.001 \\
    Weight Decay & 1e-5 \\
    Activation Function & ReLU \\
    
    Training Epochs & 1000 \\
    Optimizer & Adam \\
    Classification Output & 2 \\
    \bottomrule
\end{tabular}
}
\end{minipage}

\end{table}

The neural network architecture is implemented with PyTorch and the Deep Graph Library (DGL), allowing for efficient processing of graph-structured data. To handle bigger molecular datasets, we use multi-GPU support via data parallelization, assuring computational economy while retaining model performance. This implementation technique enables scalable processing of molecular graphs while maintaining the model's capacity to capture complicated structural patterns.

\subsection{Model evaluation}\label{metr}
To evaluate the effectiveness of our classification models, we used a range of performance indicators. Accuracy (ACC) measures the overall proportion of properly predicted cases. The Area Under the Receiver Operating Characteristic Curve (AUC) measures the model's discriminative power, indicating how well it performs compared to random guessing. Sensitivity (SE) and specificity (SP) assess the model's ability to accurately identify positive and negative examples, respectively, providing information about how well the model distinguishes between classes. In addition, the Matthews Correlation Coefficient (MCC) and the F1 Score provide a more nuanced evaluation, with MCC analyzing the balance of true and incorrect predictions across both classes and the F1 Score capturing the balance of precision and recall. The mathematical definitions of these metrics are given below.

\[
\text{ACC} = \frac{\text{TP} + \text{TN}}{\text{TP} + \text{TN} + \text{FP} + \text{FN}} 
\]


\[
\text{SE} = \frac{\text{TP}}{\text{TP} + \text{FN}} 
\]

\[
\text{SP} = \frac{\text{TN}}{\text{TN} + \text{FP}} 
\]

\[
\text{F1 Score} = \frac{2 \times \text{Precision} \times \text{Recall}}{\text{Precision} + \text{Recall}} 
\]

\[
\text{MCC} = \frac{(\text{TP} \times \text{TN}) - (\text{FP} \times \text{FN})}{\sqrt{(\text{TP} + \text{FN}) \times (\text{TP} + \text{FP}) \times (\text{TN} + \text{FN}) \times (\text{TN} + \text{FP})}} 
\]

\section{Results and discussion }\label{res}

We conducted our experiments using the widely recognised Hansen et al.\cite{hansen2009benchmark} benchmark dataset for Ames mutagenicity prediction. The final dataset consisted of 6,277 compounds after extensive pre-processing. To ensure robust model evaluation, we implemented a systematic data partitioning strategy: 80\% of the data was allocated for model development (training and validation), while the remaining 20\% was reserved for independent testing. For the development phase, we used 10-fold cross-validation on the training portion to rigorously assess model performance. Each molecular feature type underwent extensive hyperparameter optimisation to identify its optimal model configuration. This methodological approach allowed us to make fair comparisons between different molecular representations while maintaining the integrity of our evaluation framework.

Our paper presents a series of novel approaches to Ames mutagenicity prediction, including 2D scattering, GGS, and advanced geometric graph neural network-based representations (MOLG\(^3\)-SAGE), which outperform existing methods and state-of-the-art models. The findings demonstrate the advantage of our multiresolution approaches for capturing structural and spatial intricacies within molecular representations.
In Table \ref{tab:performance_comparison}, the 2D scattering image features analyzed with a Lightgbm model outperform standard chemical descriptors such as RDKit2D, Mordred, and EPCF6, as reported in \cite{van2024ampred}. The best accuracy (ACC) and area under the ROC curve (AUC) obtained with these descriptors and multiple machine learning models are 0.823 and 0.895 for Lightgbm with Mordred, 0.827 and 0.901 for Lightgbm with RDKit2D, and 0.821 and 0.896 for CatBoost with MACCS keys.

The Lightgbm model, which used 2D scattering image coefficients, outperformed the standard descriptor-based techniques, with an ACC of 0.8337 and an AUC of 0.9128. This improved performance demonstrates the benefits of 2D scattering for molecular image analysis, as the scattering features can capture finer structural details, spatial relationships and edge orientations that go beyond simple chemical descriptors, allowing the model to extract better the rich patterns and characteristics present in molecular structures.

\begin{table*}[!ht]
\centering
\caption{Comparison of the best performance of ML algorithms with different feature types and our proposed models on a test set.}
\begin{tabular}{llllllll}
\hline
Model & Feature & ACC & AUC & F1 & MCC & SE & SP \\
\hline

Lightgbm & RDKit2D & 0.827 & 0.901 & 0.843 & 0.652 & 0.860 & 0.789 \\
AMPred-CNN & image feature & 0.899 & 0.954 & 0.897 & 0.803 & 0.848 & \textbf{0.954} \\

Lightgbm &2D Image Scattering Features  & 0.8337 & 0.9128 & 0.8300 &0.6639 & 0.8125& 0.8547\\
SVC & GGS features  & 0.8382 & 0.9100 & 0.8346 & 0.6771 & 0.8166&  0.8598\\
RandomForest & GGS features  & 0.8621 & 0.9341 & 0.8585 & 0.7251 & 0.8367& 0.8875\\
XGBoost & GGS features  & 0.8775 & 0.9469 & 0.8753 & 0.7555 & 0.8598 & 0.8952\\
Lightgbm & GGS features  & 0.8767 & 0.9507 & 0.8750 & 0.7538 & 0.8629 & 0.8906\\

SVC & GGS features+GIN embeddings  & 0.9132 & 0.9747 & 0.9199 & 0.8242 & 0.9271& 0.8970\\
RandomForest & GGS features+GIN embeddings  & \textbf{0.9293} & \textbf{0.9796} & \textbf{0.9348} & \textbf{0.8589} & \textbf{0.9429}& \textbf{0.9136} \\
XGBoost & GGS features+GIN embeddings  & \textbf{0.9247} & \textbf{0.9748} & \textbf{0.9306} & \textbf{0.8495} & 0.9386& \textbf{0.9086}\\
Lightgbm & GGS features+GIN embeddings  & 0.9239 & \textbf{0.9812} & 0.9300 & 0.8470 & \textbf{0.9400}& 0.9053 \\
MOLG\(^3\)-SAGE & GGS node features  & \textbf{0.9301} & 0.9622& \textbf{0.9356}& \textbf{0.8603}&\textbf{0.9440}&\textbf{0.9136}
\\
\hline
\end{tabular}
\label{tab:performance_comparison}
\end{table*}
\begin{figure*}[!t]
    \centering
    \includegraphics[width=0.70\textwidth]{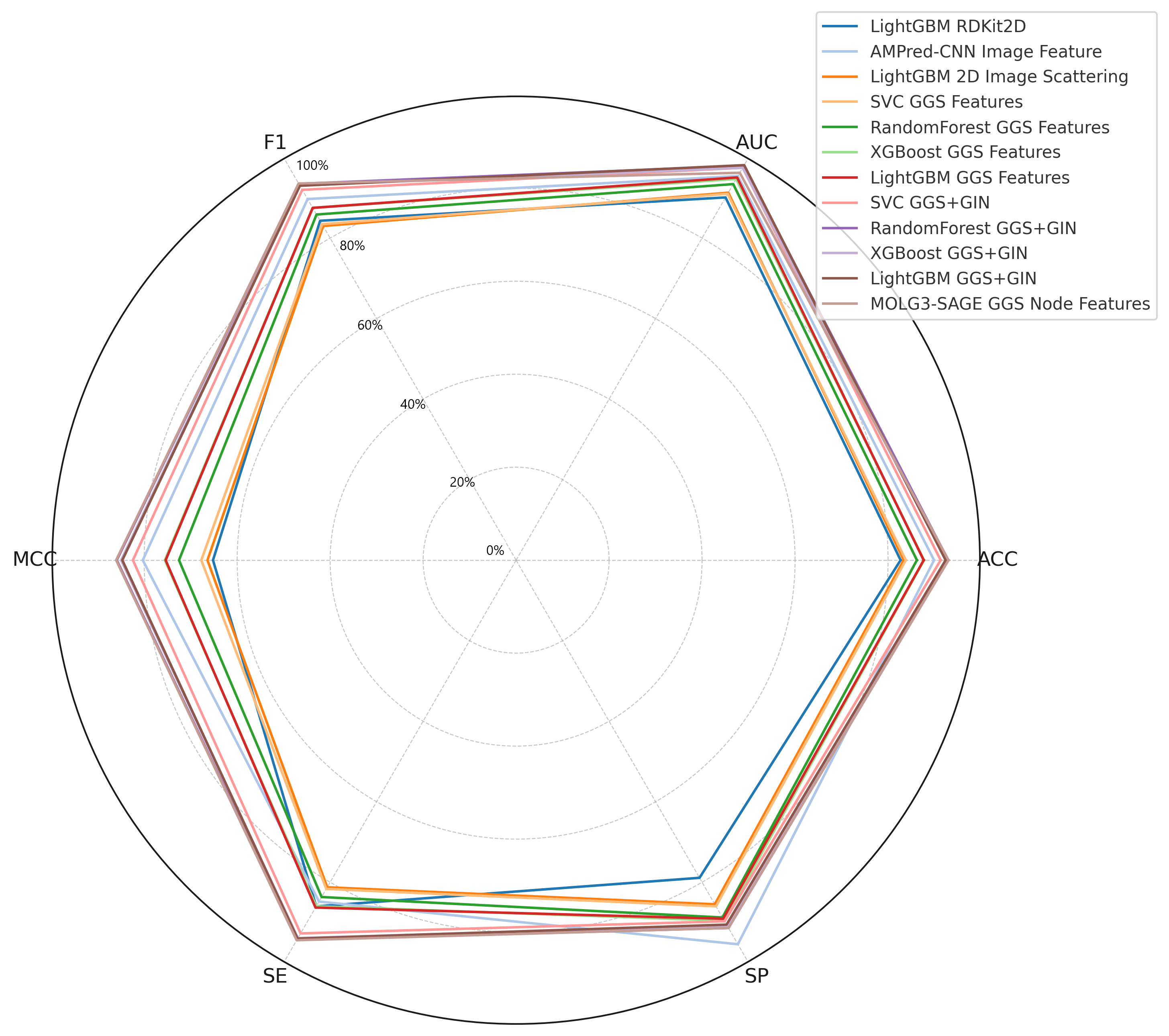}
    \vspace{-1em}  
    \caption{Comparative Performance Analysis of Machine Learning Models for Ames Mutagenecity Using Multiple Evaluation Metrics}
    \vspace{-0.5em}  
    \label{mor}
\end{figure*}
Beyond 2D scattering, we used GGS characteristics to represent molecular graphs, allowing for a more subtle and detailed representation of molecular structure and spatial configuration. Table \ref{tab:performance_comparison} and Figure \ref{mor} shows that models trained using GGS features (SVC, RandomForest, XGBoost, and Lightgbm) perform well even without the usage of deep learning architectures such as CNNs. Notably, the Lightgbm model using only GGS features achieved an AUC of 0.9507, which is very similar to the AMPred-CNN's AUC of 0.954 and more better than \cite{shinada2022optimizing,banerjee2018protox,winter2019learning} as shown in Table \ref{comp2} , demonstrating the power of GGS features in capturing structural aspects related to mutagenicity. Similarly, the XGBoost model with GGS features achieved an AUC of 0.9469, demonstrating that our graph scattering technique, even without GNN or CNN training, is a potent alternative to the cutting-edge AMPred-CNN \cite{van2024ampred}. This discovery underscores the effectiveness of GGS features in capturing important mutagenesis trends, providing comparable prediction value while requiring less computational complexity.

To increase representation capacity, we combined GGS features with node embeddings produced by a GIN. The GIN architecture employs three sequential graph isomorphism layers, each of which enriches the atomic representations (atomic number, formal charge, hybridization state, total number of hydrogens, aromaticity (binary), atomic mass, and total valence) by aggregating information from surrounding atoms.

The core of the embedding generation is the GIN architecture, which consists of three GIN convolution layers followed by linear transformations. Each GIN convolution layer employs a multi-layer perceptron (MLP) with two linear layers separated by ReLU activation functions. This architecture generates final molecular embeddings (dimension: $128$) by processing the input features (dimension: $7$) through $64$ hidden layers. The GIN layers are particularly effective at capturing structural information due to their ability to discriminate between different graph structures, making them well-suited for molecular representation learning. This hierarchical processing enables the network to learn more complex chemical patterns and structural motifs. The model's capacity to distinguish between diverse local environments while remaining permutation-invariant is a key advantage for molecular representation learning.

Following the graph convolution processes, a global pooling mechanism converts the atomic characteristics into a single molecular descriptor. This fixed-dimensional representation is further enhanced with completely connected layers, yielding 128-dimensional molecule embeddings. These embeddings capture both atomic and topological information, resulting in a comprehensive vectorial representation of the molecule structure that can be used for a variety of downstream chemical informatics activities.

\begin{figure*}[th!]
    \centering
    \includegraphics[width=0.73\textwidth]{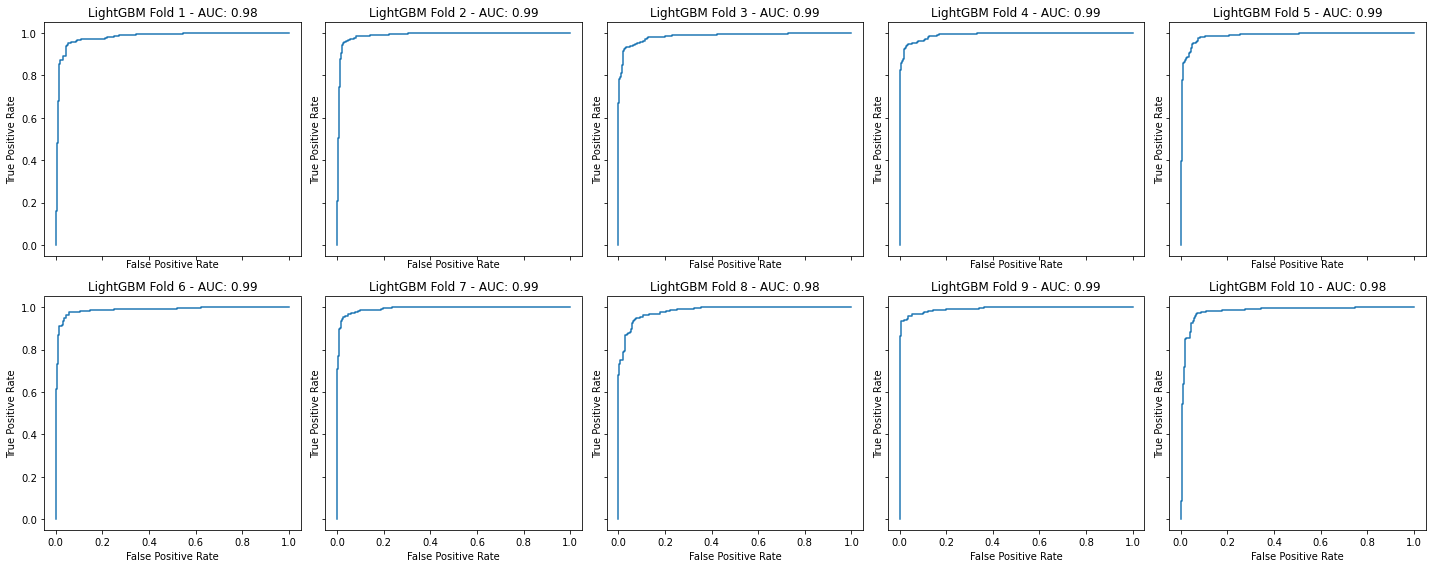}\\[5pt]
    \includegraphics[width=0.73\textwidth]{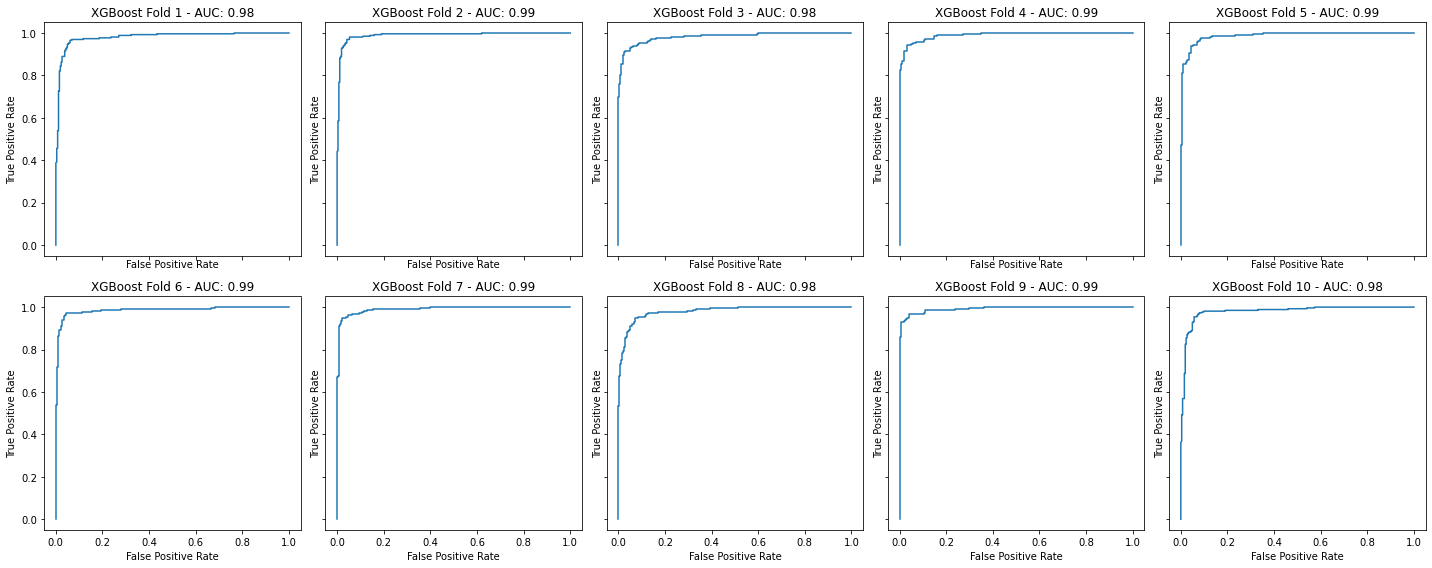}\\[5pt]
    \includegraphics[width=0.73\textwidth]{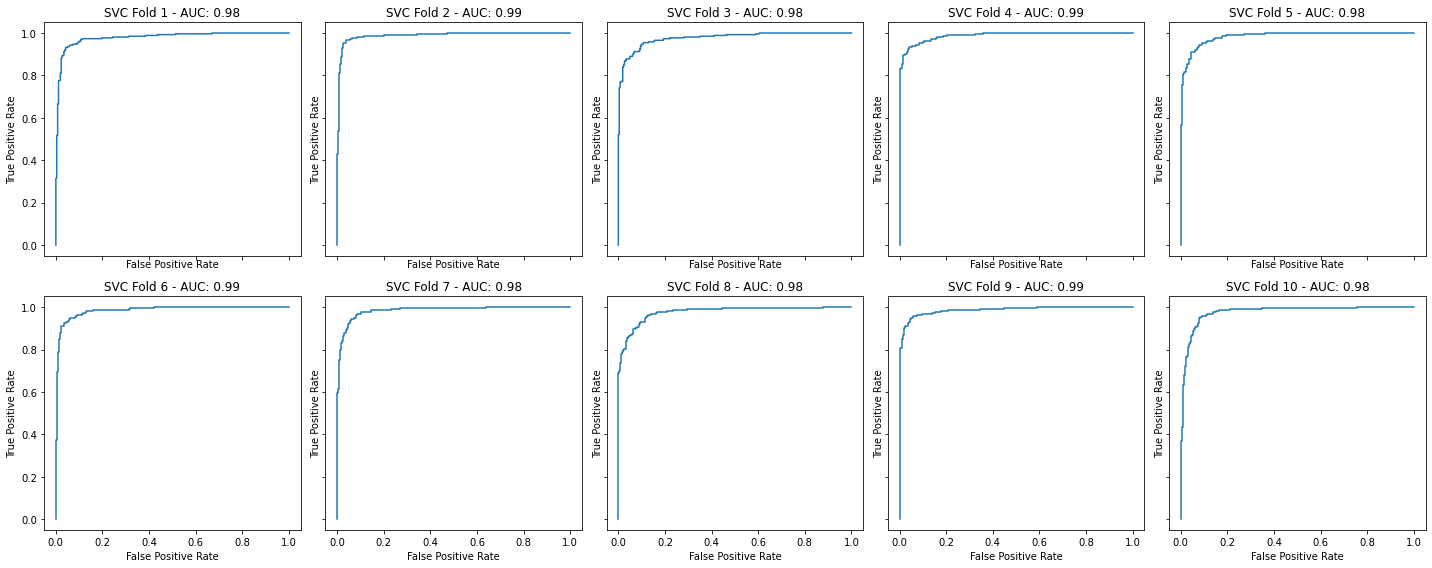}\\[5pt]
    \includegraphics[width=0.73\textwidth]{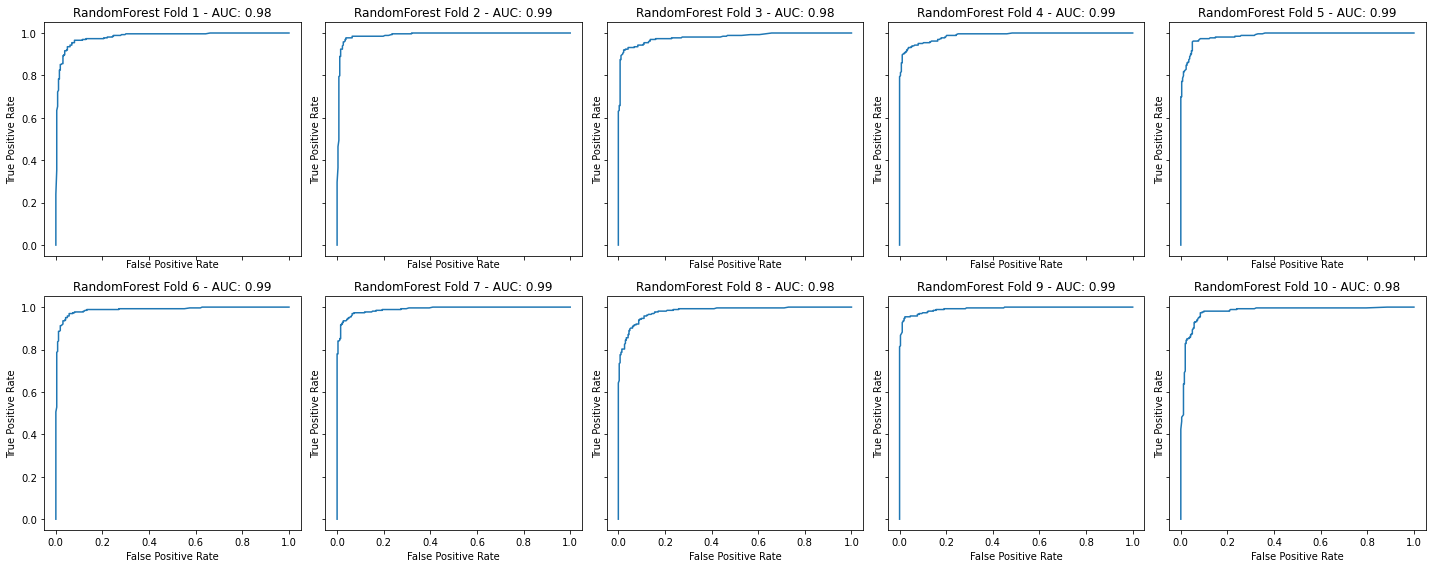}
    \caption{ROC Curves for Mutagenicity Prediction Using GGS-GIN Embeddings: Performance Comparison of RandomForest, SVC, XGBoost, and Lightgbm Models Across 10-Fold Cross-Validation}
    \label{cross}
\end{figure*}

Using the Lightgbm algorithm, this hybrid featurization achieved even better results: a model accuracy of 0.9239, an AUC of 0.9812, and an MCC of 0.8470. This extensive study of different machine learning techniques, such as Gradient Boosting, Random Forests, and Support Vector Classifiers, confirms the GGS and GIN-augmented representations' superior predictive performance. Adding to that, from 10-fold cross-validation experiments as shown in figure \ref{cross}, all models exhibit very consistent and remarkable performance, with AUC scores ranging from 0.98 to 0.99 across all folds, as evidenced by the significant rise in the ROC curves along the y-axis and proximity to the top-left corner. The  same curve shapes and AUC values across all models and folds indicate that the GGS-GIN embedding approach delivers highly discriminative features for mutagenicity prediction, resulting in robust and steady performance regardless of the classifier used.

The  MOLG\(^3\)-SAGE model with GGS node features offers a significant advance in mutagenicity prediction (figure \ref{mor}). Using GraphSAGE and GGS node features, this model obtains the best performance across practically all measures, with an ACC of 0.9301, SP of 0.9136,SE of 0.9440 and MCC of 0.8603. This configuration outperforms not only the AMPred-CNN, but also other sophisticated models that employ RNN-derived continuous vectors, MACCS, ECFP4 and other molecular properties as embeddings as illustrated in Table\ref{comp2}.

MOLG\(^3\)-SAGE uses a fully connected graph and GGS characteristics to capture complex structural and relational patterns important for predicting mutagenicity. Each node benefits from the aggregated knowledge of the complete graph, allowing the model to distinguish between mutagenic and non-mutagenic chemicals more accurately than isolated representations.
\begin{table}[!t]
\centering
\caption{Performance Comparison of AI-based Approaches utilizing Hansen et al. benchmark for Ames Mutagenicity Prediction.}
\begin{tabular}{p{0.18\textwidth}p{0.12\textwidth}p{0.16\textwidth}p{0.24\textwidth}c}
\toprule
\textbf{Approach name} & \textbf{Ref.} & \textbf{Model} & \textbf{Features} & \textbf{Reported AUC} \\
\midrule
Hansen et al. & \cite{hansen2009benchmark} & SVM & Dragon's molecular descriptor & 0.86 \\
Efficient Toxicity Prediction & \cite{karim2019efficient} & Shallow Neural Network & Padel's 2D Molecular descriptors & 0.86 \\
MutagenPred-GCNNs & \cite{li2021mutagenpred} & GCNN & Molecular graph & 0.88 \\
Rao et al. & \cite{rao2022quantitative} & GAT & Molecular graph & 0.88 \\
Winter et al. & \cite{winter2019learning} & SVM & RNN-derived continuous vectors & 0.89 \\
ProTox II & \cite{banerjee2018protox} & RF & MACCS/ECFP & 0.90 \\
Shinada et al. & \cite{shinada2022optimizing} & SVM & ECFP4, molecular properties, genotoxicity alerts & 0.93 \\
Thi et al. & \cite{van2024ampred} & Lightgbm & RDKit 2D & 0.901 \\
Thi et al. & \cite{van2024ampred} & Ampred-CNN & Image feature & 0.954 \\
\midrule
\multirow{4}{*}{Proposed Models} & \multirow{4}{*}{(Ours)} & Lightgbm & 2D Image Scattering Features & 0.9128 \\

& & Lightgbm & GGS Features & 0.9507 \\
& & MOLG\textsuperscript{3}-SAGE & Meta-Graph Geometric Scattering Features & 0.9622 \\
& & Lightgbm & GGS features, GIN embeddings & \textbf{0.9812} \\
\bottomrule
\label{comp2}
\end{tabular}
\end{table}

Finally, as shown in Table \ref{comp2}, our suggested models, particularly the Lightgbm architecture with GGS features and GIN embeddings, produce an outstanding AUC of 0.9812, exceeding previous techniques across architectures. This includes traditional SVM-based methods (Hansen et al., AUC=0.86; Winter et al., AUC=0.89; Shinada et al., AUC=0.93), neural network approaches (Efficient Toxicity Prediction, AUC=0.86), graph-based methods (MutagenPred-GCNNs, AUC=0.88; Rao et al.'s GAT, AUC=0.88), and more recent deep learning approaches like Ampred-CNN (AUC=0.954).The robust performance across many feature representations (2D Image Scattering, CGS Features, and Meta-Graph Geometric Scattering Features) proves our methodology's versatility and effectiveness. These findings set a new standard in computational toxicology, implying that our methodology may be applicable to a broader range of molecular property prediction applications. We hope that this methodology will help to improve the accuracy and efficiency of chemical safety evaluations, aiding both drug discovery and environmental safety assessments.

\section{Conclusion }\label{conc}
In this paper, we have presented a comprehensive framework for molecular mutagenicity prediction that employs a variety of methods, including geometric graph scattering , 2D image scattering, and novel graph neural network designs MOLG\textsuperscript{3}-SAGE. Our multifaceted approach considerably increases the state of the art by making numerous important contributions. First, we showed that 2D Image Scattering Features outperformed traditional molecular descriptors, including RDKit2D, MACCS, Mordred, and ECFP6, with an AUC of 0.9128 using Lightgbm. This increase confirms the efficacy of scattering transforms in extracting relevant molecular information from 2D images. More significantly, our GGS features outperformed standard descriptors, CNN, GNN, and RNN-based features in terms of capturing rich molecular structure information, resulting in remarkable prediction performance. The integration of GGS features with GIN embeddings considerably increased model performance, obtaining exceptional results with an AUC of 0.9812. Furthermore, our unique MOLG\textsuperscript{3}-SAGE design, which makes use of GGS node features, displayed exceptional performance (ACC: 0.9301, AUC: 0.9622, MCC: 0.8603), setting a new standard in the field. These findings demonstrate the efficacy of our multi-perspective strategy, which combines the advantages of scattering transforms in both 2D and graph domains with sophisticated graph neural network topologies. The higher performance across numerous feature types, machine learning techniques, and assessment criteria demonstrates the robustness and generalizability of our methods. Looking ahead, we believe our methodology has extensive implications beyond mutagenicity prediction and might be used to various molecular property prediction tasks in drug discovery and chemical safety evaluation. Future research could look into the interpretability of our models and their application to additional toxicological endpoints, perhaps leading to more efficient and accurate computational tools for chemical safety evaluation.

\bibliography{main}

\end{sloppypar}
\end{document}